\documentclass[runningheads]{llncs}

\usepackage[T1]{fontenc}
\usepackage{amsthm}
\usepackage{float}
\usepackage{algorithm}
\usepackage{algorithmic, amsmath, amssymb}
\usepackage{hyperref}
\usepackage{graphicx, tabto, subcaption, caption}
\usepackage{xcolor}
\usepackage{mdframed}

\newtheoremstyle{custom}
  {6pt} 
  {6pt} 
  {\itshape} 
  {2pt} 
  {\bfseries} 
  {.} 
  {1em} 
  {} 
\theoremstyle{custom}
\begin{document}
\title{Conformal uncertainty quantification to evaluate predictive fairness of foundation AI model for skin lesion classes across patient demographics}
%
\titlerunning{Conformal Uncertainty Quantification for Fair Skin Lesion Classification}
%
\author{Swarnava Bhattacharyya\inst{1} \and Umapada Pal\inst{1} \and Tapabrata Chakraborti\inst{2*}}
\authorrunning{S. Bhattacharyya et al.}
\institute{Indian Statistical Institute, Kolkata, India \and The Alan Turing Institute and University College London, London \\
* corresponding author: \texttt{tchakraborty@turing.ac.uk; t.chakraborty@ucl.ac.uk}
}
\maketitle              
\begin{abstract}

Deep learning based diagnostic AI systems based on medical images are starting to provide similar performance as human experts. However these data hungry complex systems are inherently black boxes and therefore slow to be adopted for high risk applications like healthcare. This problem of lack of transparency is exacerbated in the case of recent large foundation models, which are trained in a self supervised manner on millions of data points to provide robust generalisation across a range of downstream tasks, but the embeddings generated from them happen through a process that is not interpretable, and hence not easily trustable for clinical applications. To address this timely issue, we deploy conformal analysis to quantify the predictive uncertainty of a vision transformer (ViT) based foundation model across patient demographics with respect to sex, age and ethnicity for the tasks of skin lesion classification using several public benchmark datasets. The significant advantage of this method is that conformal analysis is method independent and it not only provides a coverage guarantee at population level but also provides an uncertainty score for each individual. We used a model-agnostic dynamic F1-score-based sampling during model training, which helped to stabilize the class imbalance and we investigate the effects on uncertainty quantification (UQ) with or without this bias mitigation step. Thus we show how this can be used as a fairness metric to evaluate the robustness of the feature embeddings of the foundation model (Google DermFoundation) and thus advance the trustworthiness and fairness of clinical AI.

\end{abstract}

\section{Introduction}

Skin cancer remains a significant global health concern, with melanoma accounting for more than 5\% of the total cancer cases diagnosed in the US and causing more than 8000 deaths in 2024, with multiple nonmelanoma cancer subtypes having largely unreported incidence counts in millions every year\footnote{\url{https://seer.cancer.gov/statfacts/html/melan.html}}, \cite{skin_cancer_review_3}. With over 8430 people estimated to die from melanoma in 2025 in the USA alone \cite{skin_cancer_facts_2}, figures from around the world highlight its escalating incidence. Among these, basal cell carcinoma (BCC) and squamous cell carcinoma (SCC) are the two most common forms, with over 5.9 million and 1.8 million cases recorded in 2017 respectively \cite{skin_cancer_review_3}, which are limited locally to the region of primary occurrence\cite{skin_cancer_review_1}. Meanwhile, melanoma is the most serious type of skin cancer due to its propensity for metastasis, with 75\% of deaths associated with skin cancer being caused by melanoma \cite{skin_cancer_review_1}. Geographical variability is also noteworthy, with regions such as Australia and New Zealand reporting high incidence rates, with opportunistic early detection being the major treatment method \cite{skin_cancer_review_2}. Furthermore, with more people diagnosed with skin cancer in the US each year than all other cancers combined, the urgency for continued research in skin cancer classification, prevention, and treatment has never been more critical \cite{skin_cancer_facts_2}.

The integration of artificial intelligence (AI) in dermatological practice has emerged as a transformative approach for skin cancer diagnosis. Modern deep learning based diagnostic systems have demonstrated performance levels comparable to human experts \cite{isic2018_cancer}. Due to the data-hungry nature of deep learning models, using additional metadata during training often helps in increasing the performance of such models significantly \cite{meta_deep}. and others. The field of deep learning based medical image analysis is currently seeing a shift from convolutional neural networks (CNNs) towards large vision transformer ViT based models \cite{foundation_patho_text}. These models are popularly referred to as foundation models as they are often train in a general purpose self-supervised manner over millions of images to generate a rich feature embeddings, which can then be fed into a tasks specific bespoke model for specific tasks. However all of these existing approaches share some drawbacks both from the model and data perspectives. State-of-the-art deep learning models are complex (CNNs have millions of trainable parameters while large ViTs may have billions) and hence inherently opaque to interpretation. But for such models to be adopted in high risk applications like healthcare, it is crucial to overcome this clinical translational bottleneck through decision transparency. On the other hand, it is important to leverage the power of these state-of-the-art foundation models, thus leading to a dichotomy. The second constraint is related to quality and quantity of data availability. There is a severe class imbalance problem persisting in most available healthcare datasets - this is the well known long tail problem in computer vision. However, in certain medical imaging tasks there can be additional bias across patient demographics with respect to sex, age or race. For example, in the case of skin cancer, there is a significant majority of caucasian patients and thus the model might have higher predictive accuracy for those patients leading to lack of algorithmic fairness.

Our work addresses both of the above challenges, that is predictive trustworthiness and fairness in skin lesion classification. Firstly, we do not shy away from using the cutting edge ViT based foundation models to achieve state-of-the-art performance (we use Google Derm Foundation model \cite{derm_foundation}), but rather demonstrate the robustness and trustworthiness of the model by rigorously quantifying the predictive uncertainty of the AI pipeline using conformal predcition based uncertainty quantification. By using a hold out calibration set of samples, conformal analysis provides a marginal coverage guarantee that the set of predicted labels in test phase (called the conformal set) will contain the true label at a user specified level of significance. Additionally, it provides a confidence bound for each individual patient, thereby increasing the trustworthiness of the system. To address the bias of class imbalance across patient ethnicity, we introduce a novel F1 dynamic custom sampler between training epochs and an ensemble-learning-based strategy on both sets of data with caucasian and asian patients. This resulted in increased robustness of predictive performance between both patient ethnic groups which was quantified with accuracy metrics as well as conformal uncertainty quantification. Though our work focuses on skin lesion classification, both the approaches (F1 based dynamic sampling and conformal prediction) are model agnostic and task agnostic statistical approach and thus can be used as a generalised framework for measuring algorithmic fairness.

\section{Methodology}

Our methodology for the skin lesion classification process and subsequent conformal uncertainty prediction is an approach that combines the power of state-of-the-art foundation models with the trustworthiness of conformal prediction based uncertainty quantification, as discussed in this Section.

\begin{figure}
    \centerline{\includegraphics[scale=0.3]{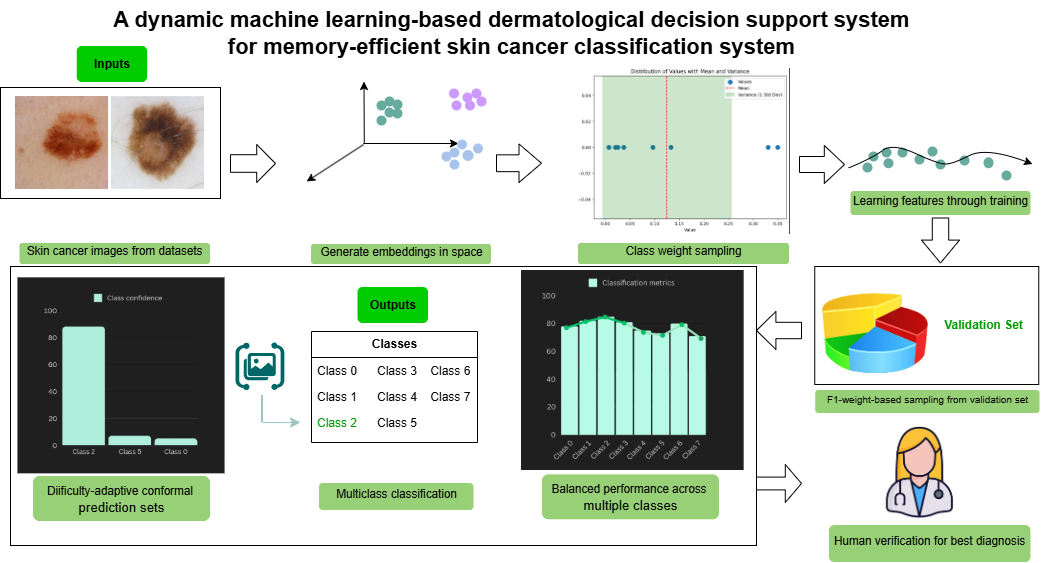}}
    \caption{Pipeline of the proposed system. Our AI-driven workflow can be integrated with skin cancer diagnosis systems for supporting manual diagnosis of patients by classifying skin cancer from dermatological images, especially in lower time and memory constraints. In real-time, it can \textbf{classify image samples into skin cancer subtypes} and \textbf{produce prediction sets showing the guarantee associated with the most-probable predictions}}
    \label{overview}
\end{figure}

\subsection{F1-weight-based dynamic sampler}
Since both datasets are heavily class imbalanced, with maximum class frequencies being several times bigger than some of the minority classes, we built a dynamic, model-agnostic epoch-wise sampling algorithm based on F1-score-based weights for classes, which greatly balanced the classwise performance. Two important parameters for our custom sampler were the \textit{threshold value \(\lambda\)} and the \textit{minimum weight \(\beta\)}, and the sampler update rule involving them is described in Algorithm \ref{alg:F!-score sampling} below. The threshold decides the cutoff for which classes will be baseline sampled and which will be F1-weight sampled, thereby maintaining a healthy balance between minority and majority classes. The minimum weight is the weight value by which the majority classes are baseline sampled. The choice of using F1-sampling over other balancing techniques is mainly its effectiveness and adaptability. Using existing resources like the validation samples, the model can effectively adjust itself periodically during the training process and focus it's learning more towards classes whose samples it is finding more challenging to classify. This balances the training schedules inherently without any external inputs. However, readjusting itself after every epoch of training might lead to overfitting the training data. This is where the adaptability of this mechanism shines --- we can deploy it as and when needed in the training pipeline based on our dataset distribution and model performance. 




\begin{algorithm}
\caption{Challenge-regulated F1-score Sampling for Highly Imbalanced Datasets}
\label{alg:F!-score sampling}
\begin{algorithmic}[1]

\REQUIRE Original dataset $D$, model in training pipeline $M$
\STATE Train $M$ on the training set of $D$ for T epochs, where T is determined from the training experiment
\STATE \textbf{F1-weight calculation strategy:} 
   \begin{itemize}
    \item Periodically, pass the validation set of $D$ through $M$ and calculate classwise F1-scores $F_s^i$. where \(i\in[0,n]\) where n is the total number of classes
   \item For each F1-score $F_s^i$, the corresponding F1-class-weight is calculated as \(F_w^i = \frac{1}{F_s^i}\)
   \item Normalize the scores in the range \((0,1)\)by division with the sum of the scores
   \end{itemize}
\STATE \textbf{Sampler update strategy}
    \begin{itemize}
    \item For each F1-class-weight $F_w^i$, if 
        \begin{equation}
            F_w^i<\lambda \Rightarrow w_i=\beta
            \label{eq:Sampler update rule 1}
        \end{equation}
        else
        \begin{equation}
        F_w^i \geq \lambda \Rightarrow w_i=F_W^i
        \label{eq:Sampler update rule 2}
        \end{equation}
    \end{itemize}
\STATE Use the updated sampler for the next T training epochs

\end{algorithmic}
\end{algorithm}

\subsection{AI model architectures}

We have used a state-of-the-art vision transformer based foundation model (Google DermFoundaiton model) for feature embedding. The advantage of using this model is that it has been specifically pre-trained on extracting robust embeddings from dermatology images, hence these embeddings can be used directly without need for fine-tuning. This enables us to use a relatively simple multi-layer perception (MLP) type neural network as the classifier head. Since the MLP network has only few hidden layers, the training overhead gets substantially reduced as the  foundation model backbone remain frozen. This keeps the model lightweight and hence suitable for deployment in clinical settings which are constrained in computational resources, while preserving the power of the foundation model. For the MLP models, we used a neural network with 2048 input neurons, 6 blocks each consisting of a fully connected layer with half the neurons from the previous block, a 1D batch normalization layer, an activation layer, and a dropout layer, followed by a final fully connected layer after the last block. This architecture proved effective with the embeddings for classification, as it correctly learned the feature representations in the embeddings. Since the embeddings generated from the two datasets were fundamentally different due to the difference in dermatological features within images, for the combined training approach, we used the Balanced Random Forest, which aggregated a large number of weak learners to produce a strong outcome based on ensemble learning. This tactic proves effective in tackling the covariate shift present in the joint distribution of the two datasets. 



\subsection{Conformal prediction for uncertainty quantification}

 Conformal prediction is a rigorous statistical calibration technique for uncertainty quantification of predictive models. At a user defined level of significance, it provides a marginal guarantee that the true prediction will be contained in the predicted set of output labels for classification or predicted range for regression tasks. Additionally, for each individual (that is test sample), it provides a uncertainty bound of prediction which is useful for personalised healthcare or precision medicine. This increases the trustworthiness of the AI predictions and the healthcare providers can make a more informed and interpretable decision based on the conformal prediction.



For our work, the steps to generate the conformal prediciton sets were as follows. We build a separate calibration set consisting of a small number of samples (approx. 500), which the model has never encountered during training or testing. We use the deviation between predicted and true labels to calculate nonconformity scores for the calibration samples, which help to define the threshold for confidence intervals. We sort the nonconformity scores and take the $1-\alpha$ quantiles with some finite correction as the threshold score for generating conformal prediction sets, where $\alpha$ is the level of significance for the coverage guarantee. 


\begin{figure}
    \centerline{\includegraphics[scale=0.5]{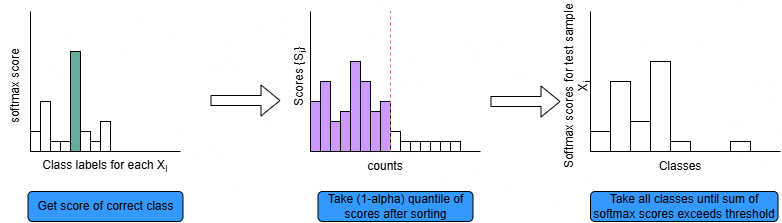}}
    \caption{Detailed steps in the conformal prediction set generation process}
    \label{fig:conformal prediction}
\end{figure}

\begin{equation}
    1-\alpha \leq \mathbb{P}\bigl(Y_{\text{test}} \in C(X_{\text{test}})\bigr)\leq 1-\alpha+\frac{1}{n+1}
    \label{eq:conformal prediction eq}
\end{equation}
where \(\bigl(X_\text{test}, Y_\text{test})\) is a test set point from the same distribution as the calibration set, and \(\alpha \in [0,1] \) is the user-chosen error rate and $n$ is the size of the calibration set. For test samples with which the model faces considerable difficulty, the number of labels in the prediction set increases --- so as to say, the length of the prediction set for any test sample is indicative of the challenge faced by the model while classifying it. 

\section{Experimental setup}


We have used two public benchmark skin lesion classification datasets for our work: the ISIC 2019 challenge dataset and the ASAN skin cancer dataset. To maintain an unbiased approach, we applied a fixed set of preprocessing transformations to images from both datasets. Each image is resized to 64x64 pixels. Random transformations, that is, horizontal and vertical flipping, rotating by 90 degrees, and transposition were employed. We adjusted the brightness and contrast of the transformed images within a set limit of 0.8 and 1.2 of the original values. The images obtained from this process were stored in a Google cloud services (GCS) bucket, from where the Derm Foundation API was used for generating the embeddings. The embeddings for each image were stored in a JSON file with the corresponding image ID as the key.



The training pipeline was built using PyTorch, and was mostly common for both datasets except for minor changes. We used a custom sampler as per our requirement, and trained each model for 40 epochs. The custom samplers were initialized with class frequency weights, calculated as: 
\begin{equation}
    w_i = \frac{\frac{1}{n_i}}{\sum_{j=1}^{C}\frac{1}{n_j}}
    \label{eq:Class frequency weights}
\end{equation}

Where \(w_i\) is the weight for class \(i\), \(\bar{x}_i\) is the mean F1 scores from \(k\) folds of cross validation for class \(i\), and \(C\) is the total number of classes. As outlined previously, the sampler was updated periodically during training to adjust the data seen by the model accordingly from the performance achieved on the validation set. During those updating processes, the following equation was used:

\begin{equation}
    w_i = \frac{\frac{1}{\bar{x}_i}}{\sum_{j=1}^{k} \frac{1}{\bar{x}_j}}
    \label{eq: F1-score-based class-weights}
\end{equation}
Where \(w_i\) is the weight for class \(i\), \(n_i\) is the sample count for class \(i\), and \(C\) is the total number of classes. Using this training pipeline, both the models were trained effectively, and the results obtained from testing with the respective test sets are discussed in the upcoming sections.

\subsection{ISIC2019 Dataset}
The ISIC 2019 Dataset combines three notable skin cancer datasets of mostly caucasian patients, viz., BCN\_20000 Dataset (by Department of Dermatology, Hospital Clínic de Barcelona), HAM10000 Dataset (by ViDIR Group, Department of Dermatology, Medical University of Vienna) and MSK Dataset. Along with 25,331 images of skin lesions, the ISIC 2019 dataset also contained additional patient metadata, like age, sex, general anatomical site, etc. The ISIC dataset comprises of the images, a CSV file containing the ground truth labels for each image, and another CSV file for the additional metadata. We took 23,254 embeddings, dropping the 'UNK'-labelled and the downsampled images as they were not suitable for classification. These images were distributed unevenly among 8 classes, viz. melanoma, melanocytic nevus, basal cell carcinoma, actinic keratosis, benign keratosis (solar lentigo/seborrheic keratosis), dermatofibroma, vascular lesion, and squamous cell carcinoma, with melanocytic nevus having the highest number of images (11,557) while dermatofibroma had the least (239). This was split into train (13,392), validation (3720), test (4654), and a special calibration set for the conformal prediction process (1488). The custom sampler for dynamic sampling in this training pipeline was initialized using class frequency weights as standard, and thereafter, the custom sampler was configured with F1-weights after every four epochs of training, calculated from the validation set. For the F1 weights, we performed 10-fold cross-validation and took the mean value of the classwise F1 scores for calculating the weights, which boosted the robustness of our algorithm. These weights were used for sampling with the custom sampler, with a threshold of one standard deviation above the mean of the provided weights, with a minimum baseline sampling of two standard deviations above the mean weight. Using this technique, we trained the MLP model for 40 epochs, followed by testing and producing conformal sets. 

\subsection{ASAN Skin Cancer Dataset}
The ASAN Dataset was built by the Department of Dermatology at the ASAN Medical Centre by collecting clinical images of skin lesions from patients of Asian demographics and annotated by dermatologists. The ASAN skin cancer dataset was introduced by Han et. al. \cite{asan_intro_paper} in their paper, where alongside the ASAN dataset, they used the MED-NODE dataset and atlas site images to build a deep learning algorithm based on the Microsoft ResNet-152 model. The ASAN dataset comprises of 12,209 images of skin disorders, viz., actinic keratosis, basal cell carcinoma, dermatofibroma, hemangioma, intraepithelial carcinoma, lentigo, melanoma, melanocytic nevus, pyogenic granuloma, squamous cell carcinoma, seborrheic keratosis, and wart.  Similar to the ISIC-2019 training pipeline, the ASAN pipeline was designed to dynamically sample instances from the data after each epoch based on F1-weights. The ASAN dataset comprised of 12,209 images, which were divided into train (8864), validation (1086), test (1274), and calibration (985) sets.  Among these classes, there is a severe imbalance, with melanocytic nevus being the largest class with 2274 instances and pyogenic granuloma being the smallest class with just 358 instances. This severe imbalance once again produced distorted results when training a classifier. We initialized the custom sampler with class frequency weights, following which, the sampler uses the F1-Score Weights to dynamically sample class instances. The sampler weights are updated with freshly-calculated F1-weights every alternate training epoch. After 40 training epochs, we test the model's performance on the test set.

\section{Results}


In this section we first present the standard performance metrics for the datasets when trained indivisually as well as together. Next, we provide an in depth set of results on the conformal prediction based uncertainty quantification with resepect to algorithmic fairness across patient demographics.  

\subsection{Classification Results}
After 40 epochs of training with F1-weights using the custom sampler, the overall accuracy on the ISIC dataset on the test set was 70.33\%, while the individual class metrics are highlighted in Table 1. The results when using the custom sampler and F1 weights during training (Sampled columns in the table) have significantly improved minority class metrics by a 3-5\% margin while maintaining unchanged performance levels for majority classes in the dataset, such as melanoma and melanocytic nevus. There is a notable jump in the F1-scores of all classes due to sampling using F1-weights, which makes the model learn more from the classes on which it is facing more difficulty in a dynamic manner during training. Under same experimental conditions, an overall accuracy of 68.83\% was obtained on the ASAN dataset; the class-wise performance metrics are provided in Table 2. Using the custom sampler and F1-weights during training helps to increase metrics by 3-5\% for minority classes compared to the unsampled training process, thereby balancing out performance metrics between classes. Thus both tables show similar trends on the 2 datasets. For our combined training approach, wherein we trained a single classifier on both datasets together to develop a more robust learning model, we achieved an overall accuracy of 65.38\% on the ASAN dataset and 72.49\% on the ISIC2019 dataset, on the six common classes of the datasets. Detailed results are provided in Table 3 and Table 4.

\begin{table}[h]
    \begin{center}
        \resizebox{\textwidth}{!}{
            \centering
            \begin{tabular}{|l|c|c|c|c|c|c|c|c|}
            \hline
            \textbf{ISIC Classes} & \textbf{Acc} & \textbf{Acc} & \textbf{F1-score} & \textbf{F1-score} & \textbf{Recall} & \textbf{Recall} & \textbf{AUC} & \textbf{AUC} \\
            & \textbf{Sampled} & \textbf{Unsampled} & \textbf{Sampled} & \textbf{Unsampled} & \textbf{Sampled} & \textbf{Unsampled} & \textbf{Sampled} & \textbf{Unsampled}\\
            \hline
            MEL & 0.6422 & 0.6181 & 0.5992 & 0.5758 & 0.6422 & 0.6181 & 0.77 & 0.59 \\ \hline
            NV & 0.7543 & 0.6951 & 0.8224 & 0.7914 & 0.7543 & 0.6951 & 0.89 & 0.91 \\ \hline
            BCC & 0.7128 & 0.6286 & 0.7281 & 0.6990 & 0.7128 & 0.6286 & 0.95 & 0.96 \\ \hline
            AK & 0.5057 & 0.5805 & 0.4665 & 0.4335 & 0.5057 & 0.5805 & 0.92 & 0.90 \\ \hline
            BKL & 0.5424 & 0.5580 & 0.5127 & 0.4669 & 0.5424 & 0.5580 & 0.81 & 0.72 \\ \hline
            DF & 0.7292 & 0.7500 & 0.5738 & 0.2562 & 0.7292 & 0.7500 & 0.95 & 0.89 \\ \hline
            VASC & 0.9412 & 0.9804 & 0.8496 & 0.6757 & 0.9412 & 0.9804 & 0.98 & 0.98 \\ \hline
            SCC & 0.7063 & 0.4365 & 0.4395 & 0.3630 & 0.7063 & 0.4365 & 0.91 & 0.89 \\ \hline
            \end{tabular}
        }
        \caption{Classwise Classification Results for ISIC 2019 Dataset}
        \label{tab:classification_metrics}
    \end{center}    
\end{table}

\begin{table}[h]
    \begin{center}
        \resizebox{\textwidth}{!}{
            \centering
            \begin{tabular}{|l|c|c|c|c|c|c|c|c|c|}
                \hline
                \textbf{ASAN Classes} & \textbf{Acc} & \textbf{Acc} & \textbf{F1-score} & \textbf{F1-score} & \textbf{Recall} & \textbf{Recall} & \textbf{AUC} & \textbf{AUC} \\
                & \textbf{Sampled} & \textbf{Unsampled} & \textbf{Sampled} & \textbf{Unsampled} & \textbf{Sampled} & \textbf{Unsampled} & \textbf{Sampled} & \textbf{Unsampled}\\
                \hline
                ak & 0.5161 & 0.6129 & 0.5120 & 0.5278 & 0.5161 & 0.6129 & 0.90 & 0.92 \\
                bcc & 0.7273 & 0.6909 & 0.6909 & 0.6756 & 0.7273 & 0.7273 & 0.93 & 0.95 \\
                dermatofibroma & 0.7500 & 0.7586 & 0.7500 & 0.7652 & 0.7500 & 0.7586 & 0.90 & 0.92 \\
                hemangioma & 0.4578 & 0.5663 & 0.5171 & 0.5411 & 0.4578 & 0.5663 & 0.89 & 0.94 \\
                Intraepithelial carcinoma & 0.4340 & 0.4151 & 0.4868 & 0.5057 & 0.4340 & 0.4151 & 0.89 & 0.84 \\
                lentigo & 0.7551 & 0.7143 & 0.7115 & 0.7778 & 0.7551 & 0.7143 & 0.95 & 0.84 \\
                melanoma & 0.8591 & 0.7455 & 0.7759 & 0.7857 & 0.8591 & 0.7455 & 0.90 & 0.95 \\
                nevus & 0.8283 & 0.8326 & 0.8126 & 0.8308 & 0.8283 & 0.8326 & 0.96 & 0.97 \\
                Pyogenic granuloma & 0.8919 & 0.5676 & 0.5641 & 0.5738 & 0.8919 & 0.5676 & 0.92 & 0.93 \\
                scc & 0.5783 & 0.5492 & 0.5564 & 0.5663 & 0.5783 & 0.5492 & 0.93 & 0.96 \\
                sebk & 0.5354 & 0.6001 & 0.5668 & 0.5742 & 0.5354 & 0.6001 & 0.74 & 0.78 \\
                wart & 0.7727 & 0.7828 & 0.7445 & 0.7579 & 0.7727 & 0.7828 & 0.94 & 0.96 \\
                \hline
            \end{tabular}
        }
        \caption{Classwise Classification Results for ASAN Datasete}
        \label{tab:asan_single_performance}
    \end{center}
\end{table}


\begin{table}[H]
    \begin{center}
        \resizebox{\textwidth}{!}{
            \centering
            \begin{tabular}{|c|c|c|c|c|c|c|c|c|}
            \hline
            \textbf{ISIC} & \textbf{Acc} & \textbf{Acc} & \textbf{F1-Score} & \textbf{F1-Score} & \textbf{Recall} & \textbf{Recall} & \textbf{AUC} & \textbf{AUC} \\
            & \textbf{sampled} & \textbf{unsampled} & \textbf{sampled} & \textbf{unsampled} & \textbf{sampled} & \textbf{unsampled} & \textbf{Sampled} & \textbf{Unsampled}
            \\ \hline
            AK   & 72.41       & 70.11         & 46.15            & 46.12              & 72.41          & 70.11            & 0.95        & 0.95          \\ \hline
            BCC  & 71.28       & 72.63         & 69.96            & 70.20              & 71.28          & 72.63            & 0.95        & 0.95          \\ \hline
            DF   & 33.33       & 31.25         & 29.09            & 28.04              & 33.33          & 31.25            & 0.94        & 0.94          \\ \hline
            MEL  & 65.42       & 66.87         & 62.63            & 62.78              & 65.42          & 66.87            & 0.84        & 0.85          \\ \hline
            NV   & 79.54       & 76.86         & 84.84            & 84.27              & 79.54          & 76.86            & 0.92        & 0.93          \\ \hline
            SCC  & 20.63       & 19.84         & 22.61            & 23.47              & 20.63          & 19.84            & 0.92        & 0.93          \\ \hline
            \end{tabular}
        }
        \caption{Performance metrics for ISIC testset}
        \label{tab:ISIC_metrics}
    \end{center}
\end{table}

\begin{table}[H]
    \begin{center}
        \resizebox{\textwidth}{!}{
            \centering
            \begin{tabular}{|l|c|c|c|c|c|c|c|c|}
            \hline
            \textbf{ASAN} & \textbf{Acc} & \textbf{Acc} & \textbf{F1-Score} & \textbf{F1-Score} & \textbf{Recall} & \textbf{Recall} & \textbf{AUC} & \textbf{AUC} \\
            & \textbf{sampled} & \textbf{unsampled} & \textbf{sampled} & \textbf{unsampled} & \textbf{sampled} & \textbf{unsampled} & \textbf{Sampled} & \textbf{Unsampled} \\
            \hline
            ak & 74.19 & 74.19 & 57.86 & 59.35 & 74.19 & 74.19 & 0.95 & 0.95 \\
            bcc & 36.36 & 43.64 & 44.20 & 50.53 & 36.36 & 43.64 & 0.88 & 0.89 \\
            dermatofibroma & 91.38 & 90.52 & 75.71 & 74.73 & 91.38 & 90.52 & 0.96 & 0.96 \\
            melanoma & 59.32 & 59.32 & 67.31 & 66.67 & 59.32 & 59.32 & 0.97 & 0.97 \\
            nevus & 64.81 & 65.67 & 73.66 & 75.00 & 64.81 & 65.67 & 0.94 & 0.94 \\
            scc & 63.93 & 67.21 & 57.68 & 61.89 & 63.93 & 67.21 & 0.90 & 0.90 \\
            \hline
            \end{tabular}
        }
        \caption{Performance metrics for ASAN testset}
        \label{tab:asan_performance}
    \end{center}
\end{table}

\subsection{Conformal Set Prediction}

\begin{figure}
  \centering
      \begin{subfigure}{0.5\textwidth}
        \includegraphics[width=\linewidth]{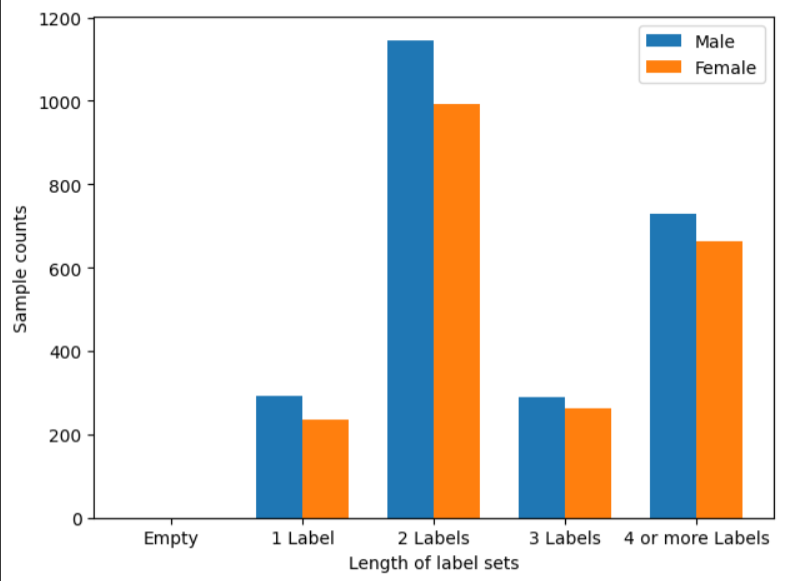}
        \captionsetup{width=0.9\linewidth}
        \caption{Difficulty associated with prediction sets variation with gender of patient}
        \label{fig:3a}
      \end{subfigure}%
      \hfill
      \begin{subfigure}{0.5\textwidth}
        \includegraphics[width=1.2\linewidth, height=0.8\textwidth]{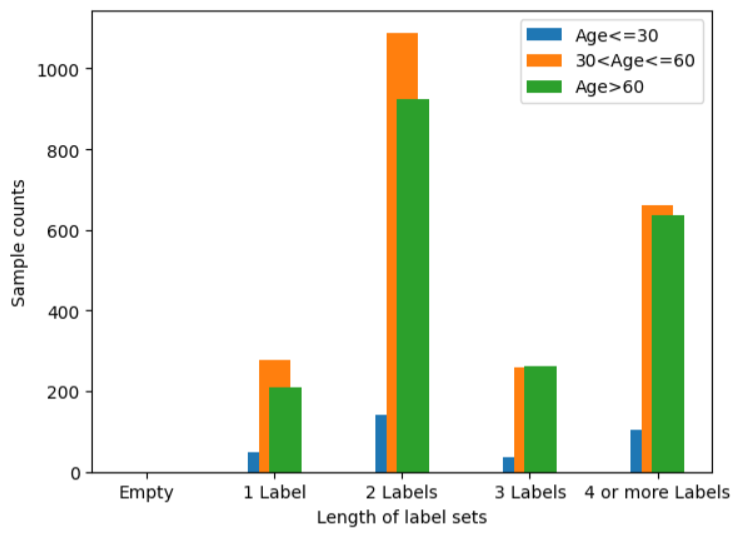}
        \caption{Difficulty associated with prediction sets variation with age of patient}
        \label{fig:3b}
      \end{subfigure}%
      \vskip\baselineskip      
      \begin{subfigure}{0.5\textwidth}
        \includegraphics[width=1.2\linewidth]{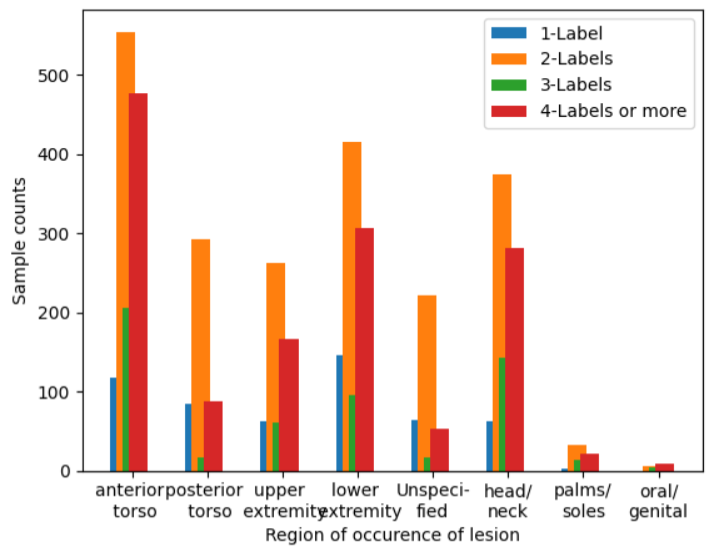}
        \caption{Difficulty associated with prediction sets variation with anatomical location of skin disorder}
        \label{fig:3c}
      \end{subfigure}%
  \caption{Variation of prediction set difficulty with patient metadata from ISIC2019}
  \label{fig:3_diffi_ISIC}
\end{figure}

For conformal prediction sets with 80\% coverage guarantee, we plotted the results for the ISIC and ASAN datasetsc respectively. From Fig. \ref{fig:3a}, we can observe that the majority of the test samples have 1 or 2 labels in their prediction sets, which displays a higher confidence of the model in these test samples. A general trend observed is that skin cancer cases are observed more in male patients than female patients. From Fig. \ref{fig:3b}, we can observe that the majority of the test set samples have 1 or 2 labels in their conformal prediction set, with the majority of the patients being spread out in the 30-60 years age range. This output, therefore, shows that model is tight confidence bounds prediction for the majority of the test samples. From Fig. \ref{fig:3c}, we can draw an important conclusion regarding the difficulty faced by the model in generating the conformal prediction sets with respect to the anatomical site of occurrence. We observe that the anterior and posterior torso are the most common spots for skin cancer occurrence, they can be classified relatively easily with the majority of sets containing 1 or 2 labels. 

\begin{figure}
  \centering
      \begin{subfigure}{0.5\textwidth}
        \includegraphics[width=\linewidth]{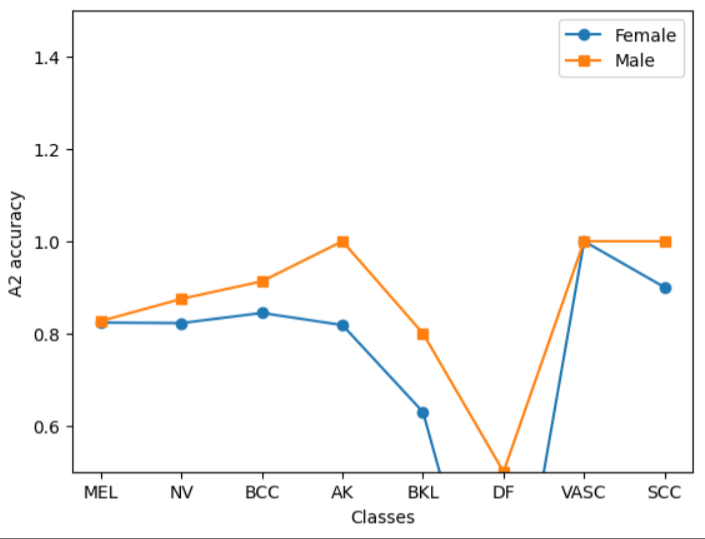}
        \captionsetup{width=0.9\linewidth}
        \caption{Classwise A2 accuracy for patients aged less than 30 years from ISIC2019 dataset}
        \label{fig:4a}
      \end{subfigure}%
      \hfill
      \begin{subfigure}{0.5\textwidth}
        \includegraphics[width=\linewidth]{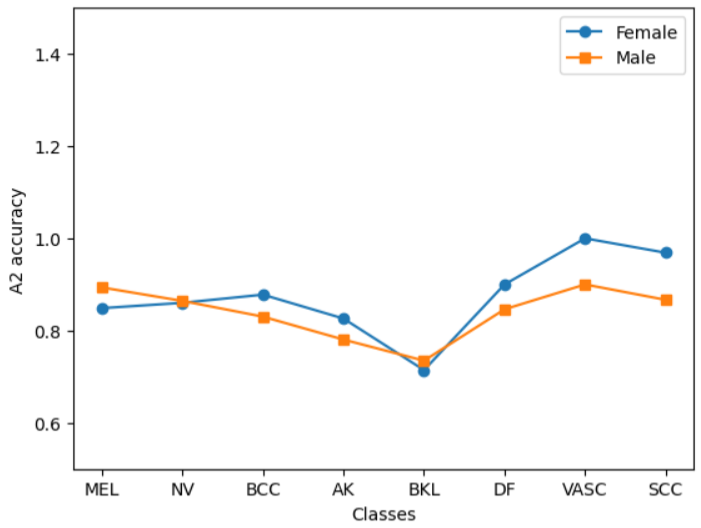}
        \captionsetup{width=0.9\linewidth}
        \caption{Classwise A2 accuracy for patient age greater than 30 years and lesser than or equal to 60 years from ISIC2019 dataset}
        \label{fig:4b}
      \end{subfigure}%
      \vskip\baselineskip      
      \begin{subfigure}{0.5\textwidth}
        \includegraphics[width=\linewidth]{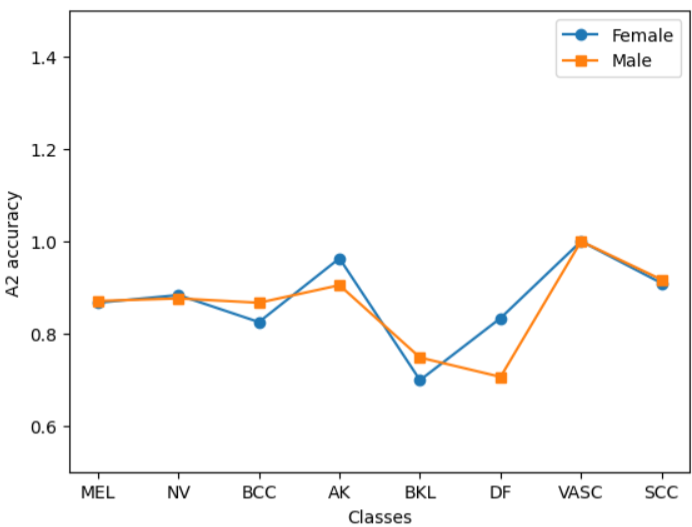}
        \captionsetup{width=0.9\linewidth}
        \caption{Classwise A2 accuracy for patient age greater than 60 years from ISIC2019 dataset}
        \label{fig:4c}
      \end{subfigure}%
      \hfill
      \begin{subfigure}{0.5\textwidth}
        \includegraphics[width=1.5\linewidth, height=0.9\textwidth]{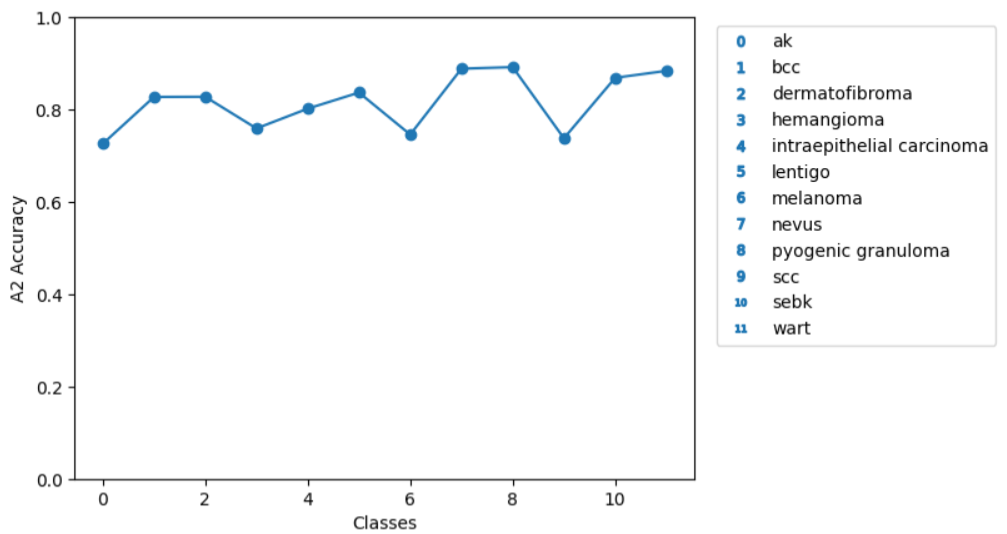}
        \captionsetup{width=0.9\linewidth}
        \caption{Classwise A2 accuracy for patients from ASAN dataset}
        \label{fig:4d}
      \end{subfigure}
  \caption{A2 accuracy for different patient categories from ISIC2019 and ASAN}
  \label{fig:4_A2}
\end{figure}

For a deeper representation, we next introduce a metric - A2 accuracy, which can be defined as the number of test samples per class that have the ground-truth label among the two most probable labels in the prediction sets, out of the total number of test samples for that class. In Figs. \ref{fig:4a}, \ref{fig:4b} and \ref{fig:4c}, we have created a graphical representation of the classwise A2 accuracy for the ISIC dataset. Fig. \ref{fig:4a} represents statistics with patients aged below 30 years; Fig. \ref{fig:4b} represents statistics with patients aged between 30 and 60 years; while Fig. \ref{fig:4c} contains statistics for patients aged over 60 years. Over both male and female patients, we can observe that A2 accuracy values lie between 80 and 100\% for all age ranges. Similarly, for the ASAN dataset, the A2 accuracy is shown in Fig. \ref{fig:4d}, where we observe a similar observation hovers around the 70 to 90\% range for all classes, which serves as a credible proof that our model provides considerably accurate coverage within the two most probable predictions. Note that ASAN dataset does not have the patient metadata with respect to age, gender and anatomical sights and hence only one sub-figure for that dataset.

\begin{figure}
  \centering
      \begin{subfigure}{0.5\textwidth}
        \includegraphics[width=\linewidth]{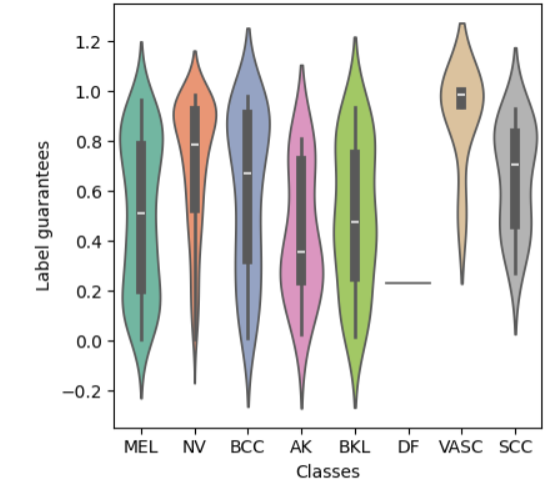}
        \captionsetup{width=0.9\linewidth}
        \caption{Classwise violin plots for patients aged less than 30 years from ISIC2019 dataset}
        \label{fig:5a}
      \end{subfigure}%
      \hfill
      \begin{subfigure}{0.5\textwidth}
        \includegraphics[width=\linewidth]{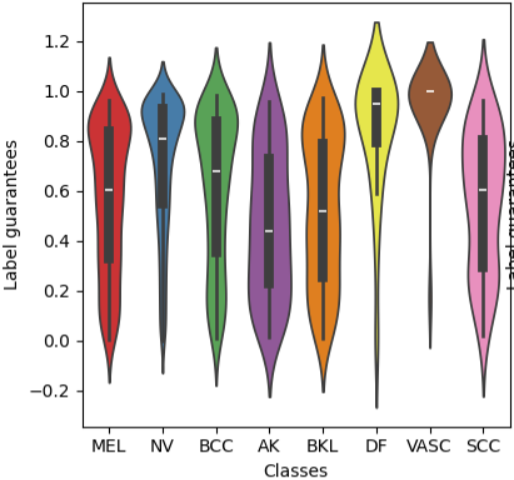}
        \captionsetup{width=0.9\linewidth}
        \caption{Classwise violin plots for patient age greater than 30 years and lesser than or equal to 60 years from ISIC2019 dataset}
        \label{fig:5b}
      \end{subfigure}%
      \vskip\baselineskip      
      \begin{subfigure}{0.5\textwidth}
        \includegraphics[width=\linewidth]{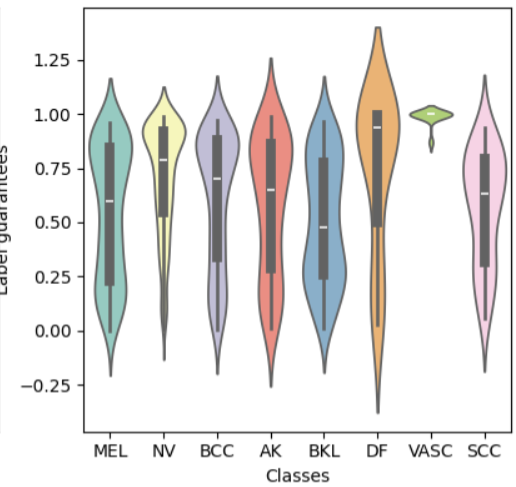}
        \captionsetup{width=0.9\linewidth}
        \caption{Classwise violin plots for patient age greater than 60 years from ISIC2019 dataset}
        \label{fig:5c}
      \end{subfigure}%
      \hfill
      \begin{subfigure}{0.5\textwidth}
        \includegraphics[width=1.5\linewidth, height=\textwidth]{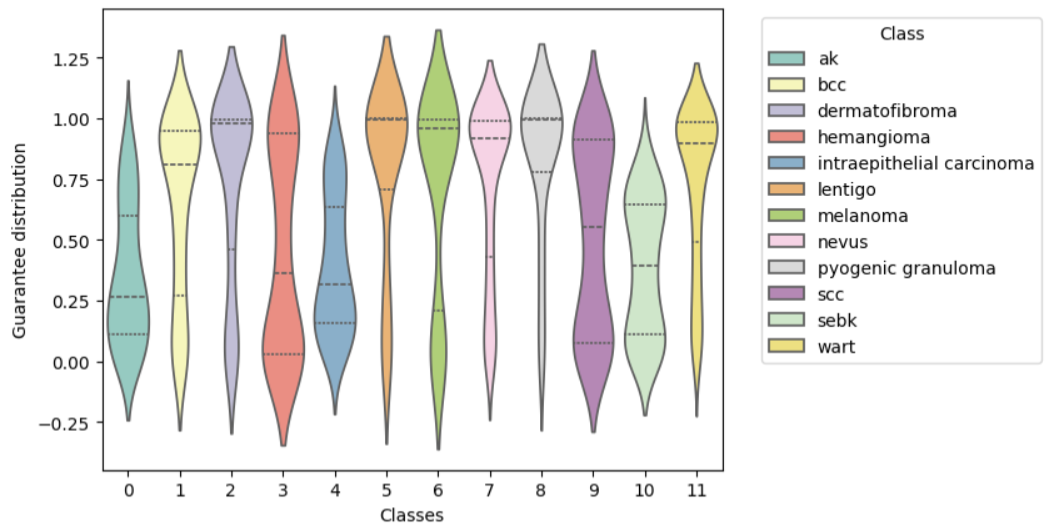}
        \captionsetup{width=0.9\linewidth}
        \caption{Classwise violin plots for patients from ASAN dataset}
        \label{fig:5d}
      \end{subfigure}
  \caption{Violin plots for different patient categories from the ISIC2019 and the ASAN datasets showing ground-truth level confidences}
  \label{fig:5_violin_plots}
\end{figure}

An alternative way of visualizing the performance of the conformal prediction pipeline is to build classwise violin plots that show the distribution of the ground-truth label confidence from each set containing it as one of the possible predictions. Essentially, we should be looking out for clusters representing unimodal distributions at the upper halves of the plot. In simpler terms, this pattern would help us conclude the model provides a guarantee in the upper half (50–100\%) for the ground-truth labels, showing considerable confidence in the correct predictions. We can see that pattern reflected in most of the violin plots, with the mode of the distribution lying closer to one.  Figs. \ref{fig:5a}, \ref{fig:5b} and \ref{fig:5c} contains three plots of classwise violin plots from the ISIC dataset, again divided by patient age; they contain patients' data with ages lower than 30 years, between 30 and 60 years, and more than 60 years, respectively. As observed, the majority of the violin plots are skewed towards one, indicating that our model gives considerably confident predictions for the ground-truth labels for test samples. For the ASAN dataset, since most prediction sets contained multiple labels, the confidence for the ground-truth label was slightly reduced, despite being among the highest ones in that prediction set, as the total confidence coverage of 80\% was distributed among multiple labels as seen from Fig. \ref{fig:5d}. For multiclass classification problems with complex features and a larger number of classes, this could be a potential issue. For the majority classes like melanoma and melanocytic nevus, due to the large number of test samples, the plots are more dense when compared to minority classes like dermatofibroma and vascular lesion. For these scatter plots, our target is to have sparse points towards zaro, which would indicate that the ground-truth confidence lies among the top two predictions in the set and has either a majority or considerable guarantee (and might require examination from a human expert to take the final decision.

\begin{figure}[H]
  \centering
      \begin{subfigure}{0.5\textwidth}
        \includegraphics[width=\linewidth, height=0.65\textwidth]{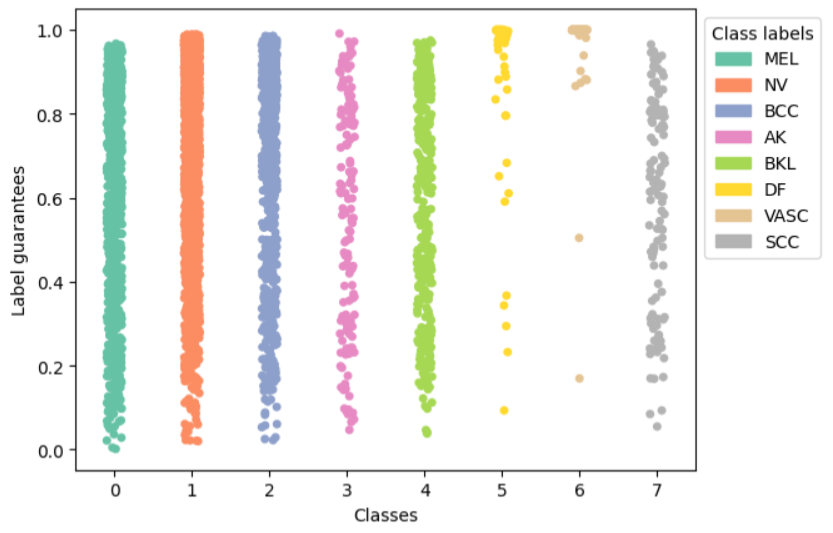}
        \captionsetup{width=0.9\linewidth}
        \caption{Classwise guarantee values for groundtruth label in top two confidences of prediction sets from ISIC2019 test samples}
        \label{fig:6a}
      \end{subfigure}%
      \begin{subfigure}{0.5\textwidth}
        \includegraphics[width=\linewidth, height=0.65\textwidth]{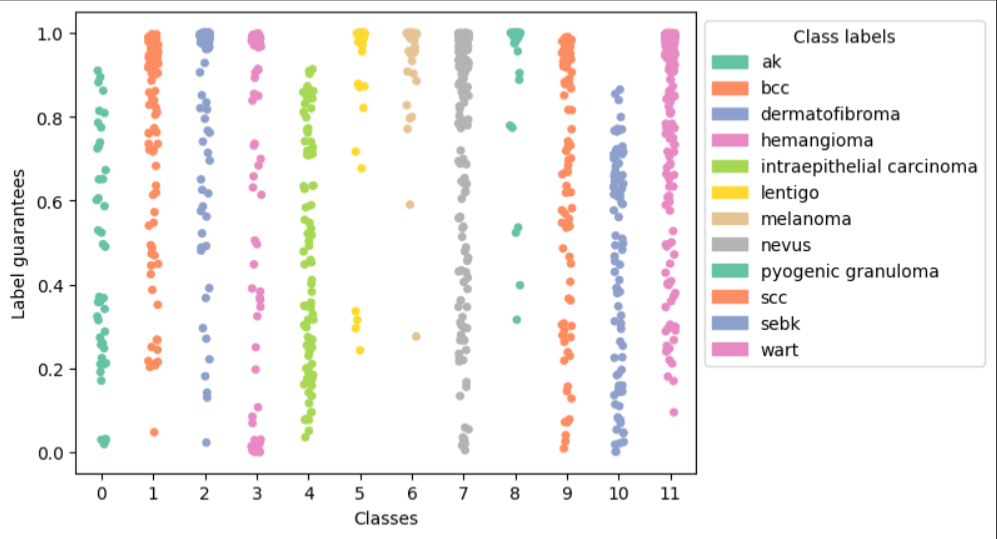}
        \captionsetup{width=0.9\linewidth}
        \caption{Classwise guarantee values for groundtruth label in top two confidences of prediction sets from ASAN test samples}
        \label{fig:6b}
      \end{subfigure}%
  \caption{Classwise guarantee values for groundtruth label in top two confidences of prediction sets from test samples}
  \label{fig:6}
\end{figure}

An effective strategy for tackling the problem of low confidence due to multiple labels can be developed by combining the strategies of the A2 accuracy and ground-truth guarantee, by observing the ground-truth guarantee if it is present in the top two guarantees of the prediction set. Fig. \ref{fig:6a} shows the scatter plot for such ground-truth guarantees, if present among the top two labels of respective prediction sets for the ISIC dataset.  All the classwise scatter plots in Fig. \ref{fig:6a} have concentrated clusters towards the higher confidence regions, and some scattered points nearer to zero. Utilising the additional metadata for the ISIC dataset, we also recorded the most common anatomical region of occurrences for skin disorders where the ground-truth lies among the top two predictions in the prediction set. This data can be of great use when using the model for diagnosis applications, as a high guarantee towards a particular class for a test sample, especially in one of the more commonly-occurring anatomical regions, can be safely considered as a correct diagnosis. This data is summarized in table \ref{tab:isic_toptwo_meta}, where the most common anatomical region of occurrences are listed in decreasing order of occurrence. For the ASAN dataset, we created a similar scatter plot with ground-truth prediction among the top two predictions of the set, as shown in Fig. \ref{fig:6b}. All the classwise scatter plots have dense clusters towards the higher guarantee regions, and only some scarce points around zero. Thus conformal prediction based uncertainty quantification when presented in different ways teases out a lot of valuable information regarding robustness and fairness of performance across different patient groupings, thus serving as a generalised metric for algorithmic fairness.

\begin{table}[h]
    \centering
    \resizebox{\textwidth}{!}{
        \centering
        \begin{tabular}{|l|c|c|c|c|c|c|c|c|}
        \hline
        \textbf{Class} & \textbf{MEL} & \textbf{NV} & \textbf{BCC} & \textbf{AK} & \textbf{BKL} & \textbf{DF} & \textbf{VASC} & \textbf{SCC} \\ \hline
        \textbf{Region 1} & Lower & Anterior & Anterior & Anterior & Anterior & Anterior & Anterior & Anterior \\ 
        &  extremity (27.34\%) & torso (25.15\%) & torso (41.12\%) & torso (42.00\%) & torso (39.94\%) & torso (39.47\%) & torso (40.00\%) & torso (38.79\%) \\ \hline
        \textbf{Region 2} & Posterior & Lower & head/ & head/ & head/& head/ & head/& head/ \\ \
        & torso (21.76\%) & extremity (19.45\%) & neck (25.13\%) & neck (19.66\%) &  neck (27.24\%) & neck (18.42\%) &  neck (22.00\%) & neck (22.41\%) \\ \hline
        \textbf{Region 3} & Anterior & Anterior & Upper & Lower & head/ & head/ & Lower & Lower \\ 
        & torso (13.67\%) & torso (17.70\%) & extremity (18.80\%) & extremity (16.88\%) & neck (19.20\%) & neck (18.42\%) & extremity (22.00\%) & extremity (20.69\%) \\ \hline
        \textbf{Region 4} & nan & Posterior & Upper & Upper & Upper & Upper & Upper & Upper \\ 
        & (13.53\%) & torso (13.05\%) & extremity (10.37\%) & extremity (12.67\%) & extremity (9.29\%) & extremity (10.53\%) & extremity (16.00\%) & extremity (15.22\%) \\ \hline
        \textbf{Region 5} & Upper & Upper & palms/ & nan & nan & palms/ & & palms/ \\ 
        & extremity (13.11\%) & extremity (12.50\%) & soles (2.64\%) & (2.00\%) & (2.79\%) & soles (5.26\%) & & soles (1.72\%) \\ \hline
        \textbf{Region 6} & head/& nan & nan & palms/ & palms/  & nan & & nan \\ 
        &  neck (10.60\%) & (10.15\%) & (1.05\%) & soles (1.33\%) & soles (1.24\%)& (5.26\%) & & (0.86\%) \\ \hline
        \textbf{Region 7} & & palms/ & oral/ & oral/ & oral/ & oral/ & & \\
        & & soles (1.65\%) & genital (0.88\%) & genital (0.67\%) & genital(0.31\%) & genital(2.63\%) & & \\ \hline
        \textbf{Region 8} & & oral/ & & & & & &\\ 
        & & genital(0.35\%) & & & & & & \\ \hline
    \end{tabular}
    }
    \caption{Classwise distribution of most common region of occurrences}
    \label{tab:isic_toptwo_meta}
\end{table}

\section{Conclusion}


In this work, we have demonstrated that conformal prediction based uncertainty quantification can function as a powerful metric for algorithmic fairness and robustness, that provides a coverage guarantee at a user specified level of significance that the true prediction is contained within the `conformal set'. This adds a level of trustability to the AI pipeline towards adoption in high risk applications like healthcare. We have chosen skin lesion classification as the predictive task in this paper because it provides a strong exemplar of class imbalance due to overwhelming ethnic bias in favour of caucasian patients. We have introduced a novel dynamic sampling strategy that uses F1 scores during training to select samples judiciously across challenging classes and hence ends up with a more equitable performance across patient demographics. Both the conformal prediction and the F1 dynamic sampler are task and model agnostic frameworks which can be generalised to other similar tasks and datasets. Despite the deluge of papers being published in health AI, very few of them get deployed to the clinic, and this clinical translation bottleneck will only get exacerbated with emerging legislations around the world regarding AI safety around the work in high risk applications like healthcare. In such a scenario, a simple yet rigorous approach like conformal prediction can help AI developers to add a layer of trustworthiness to their model without having to compromise on the deep learning architecture itself. Finally, since our method provides a bespoke conformal set for each individual patient, it can also be a progressive step towards the grand challenge of personalised healthcare and precision medicine. 

\section*{Author statement}

Authors declare no conflict of interest. T Chakraborti is supported by the Turing-Roche Strategic Partnership.

\end{document}